\documentclass[10pt,twocolumn,letterpaper]{article}

\usepackage{cvpr}
\usepackage{times}
\usepackage{epsfig}
\usepackage{graphicx}
\usepackage{amsmath}
\usepackage{amssymb}
\usepackage{ulem}
\usepackage{hhline}
\usepackage{array,booktabs,calc}
\usepackage{placeins}
\usepackage{enumitem}
\usepackage{graphicx}
\usepackage{multirow}
\usepackage{wrapfig}
\usepackage{paralist}
\usepackage{float}
\usepackage{colortbl}
\usepackage{bbm}
\usepackage{subfigure}
\usepackage{makecell}

\usepackage[pagebackref=true,breaklinks=true,letterpaper=true,colorlinks,bookmarks=false]{hyperref}

\cvprfinalcopy %

\def\cvprPaperID{2674} %

\ifcvprfinal\fi
\begin{document}

\title{Disp R-CNN: Stereo 3D Object Detection via Shape Prior Guided \\ Instance Disparity Estimation}

\author{
    Jiaming Sun$^{1,2}$ 
    \thanks{The first two authors contributed equally. The authors from Zhejiang University are affiliated with the State Key Lab of CAD\&CG.}
    \quad Linghao Chen$^{1*}$ 
    \quad Yiming Xie$^{1}$ 
    \quad Siyu Zhang$^{3}$ \\
    \quad Qinhong Jiang$^{2}$ 
    \quad Xiaowei Zhou$^{1}$\thanks{Corresponding authors: Hujun Bao and Xiaowei Zhou.}
    \quad Hujun Bao$^{1\dag}$ \\
    $^1$Zhejiang University \quad 
    $^2$SenseTime \quad
    $^3$Southern University of Science and Technology \\
}
\maketitle
\begin{abstract}
  
In this paper, we propose a novel system named Disp R-CNN for 3D object detection from stereo images.
Many recent works solve this problem by first recovering a point cloud with disparity estimation and then apply a 3D detector. The disparity map is computed for the entire image, which is costly and fails to leverage category-specific prior. 
In contrast, we design an instance disparity estimation network (iDispNet) that predicts disparity only for pixels on objects of interest and learns a category-specific shape prior for more accurate disparity estimation.
To address the challenge from scarcity of disparity annotation in training, we propose to use a statistical shape model to generate dense disparity pseudo-ground-truth without the need of LiDAR point clouds, which makes our system more widely applicable.
Experiments on the KITTI dataset show that, even when LiDAR ground-truth is not available at training time, Disp R-CNN achieves competitive performance and outperforms previous state-of-the-art methods by 20\% in terms of average precision.
The code will be available at \url{https://github.com/zju3dv/disprcnn}.

\end{abstract}

\section{Introduction}\label{sec:intro}
3D object detection plays an important role in many applications such as autonomous driving and augmented reality.
While most methods work with the LiDAR point cloud as input, stereo image-based methods have significant advantages. 
RGB images provide denser and richer color information compared to the sparse LiDAR point clouds while requiring a very low sensor price.
Stereo cameras are also able to perceive longer distances with customizable baseline settings.
Recently, learning-based approaches like \cite{kendall2017end,changPyramidStereoMatching2018,zhang2019ga} tackled the stereo correspondence matching problem with Convolutional Neural Networks (CNNs) and achieved impressive results.
Taking an estimated disparity map 
as the input,
3D object detection methods \cite{xu2018multi,wang2019pseudo} convert it into a depth map or a point cloud to detect objects within it.
However, since the disparity estimation network is designed for general stereo matching instead of the 3D object detection task, these pipelines have two major drawbacks.
First, the disparity estimation process operates on the full image and often fails to produce accurate disparities on low textured or non-Lamberterian surfaces like the surface of vehicles, which are exactly the regions we need to do successful 3D bounding boxes estimation.
Moreover, since foreground objects of interest usually occupy much fewer space than the background in the image,
the disparity estimation network and the 3D detector spend a lot of computation on regions that are not needed for object detection and lead to a slow running speed.
\begin{figure}[tb]
    \centering
    \vspace{-5pt}
    \includegraphics[width = 1.0\linewidth]{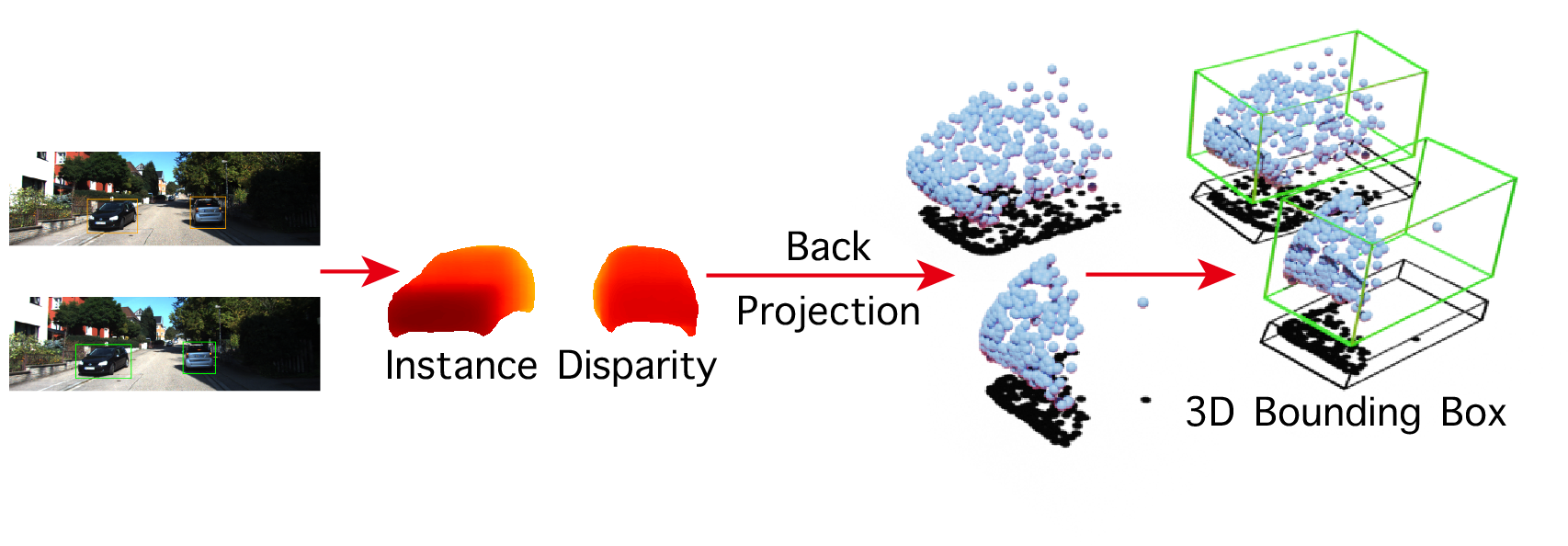}
    \caption{
        \textbf{The proposed system estimates an instance disparity map,} i.e., pixel-wise disparities only on foreground objects, for stereo 3D object detection. This design leads to better disparity estimation accuracy and faster run-time speed.
    }
    \vspace{-10pt}
    \label{fig:overview}
\end{figure}

In this work, we aim to explore how we can solve these drawbacks with a disparity estimation module that is specialized for 3D object detection.
We argue that estimating disparities on the full image is suboptimal in terms of network feature learning and runtime efficiency.
To this end, we propose a novel system named Disp R-CNN that detects 3D objects with a network designed for instance-level disparity estimation.
The disparity estimation is performed only on regions that contain objects of interest,
thus enabling the network to focus on foreground objects and learn a category-specific shape prior that is suitable for 3D object detection.
As demonstrated in the experiments, with the guidance of object shape prior, the estimated instance disparities capture the smooth shape and sharp edges of object boundaries while being more accurate than the full-frame counterpart.
With the design of instance-level disparity estimation, the running time of the overall 3D detection pipeline is reduced thanks to the smaller number of input and output pixels and the reduced range of cost volume search in the disparity estimation process.

Another limitation of the full-frame disparity estimation is the lack of pixel-wise ground-truth annotation.
In the KITTI dataset \cite{geigerAreWeReady2012} for example,
although it is possible to render disparity ground truth by manually selecting and aligning vehicle CAD models as in the KITTI Scene Flow benchmark \cite{menzeObjectSceneFlow2015}, 
there is no such ground-truth provided in the KITTI Object Detection benchmark due to its difficulty in annotating on a massive scale.
To make dense instance-level disparity supervision possible,
we propose a pseudo-ground-truth generation process that can acquire accurate instance disparities and instance segmentation masks via object shape reconstruction and rendering.
The object mesh is reconstructed by a PCA-based statistical shape model under several geometric constraints \cite{leventon2002statistical, engelmann2016joint}.
The effort to manually annotate CAD models can be saved through this automated process since the basis of the statistical shape model can be learned directly from 3D model repositories like ShapeNet \cite{chang2015shapenet}.
Different from some recent methods \cite{wang2019pseudo,you2019pseudo,chen2020dsgn} that use the projected LiDAR point clouds as the sparse supervision for full-frame disparity estimation, our pseudo-ground-truth generation process can provide dense supervision even when LiDAR is not available at training time, which has a broader applicability in practice.

We evaluate our system on the KITTI dataset and provide ablation analysis of the different components of the proposed system.
The experiments show that, with the guidance of the shape prior introduced by both the network design and the generated pseudo-ground-truth, the performance of instance-level disparity estimation surpasses the full-frame counterpart by a large margin.
As a result, 3D object detection performance can be largely improved compared to baseline state-of-the-art 3D detectors that rely on full-frame disparities.
When LiDAR supervision is not used at training time, our method outperforms the baseline methods by 20\% in terms of average precision (27\% vs. 47\%).

In summary, our contributions are as follows:
\begin{compactitem}
 \item A novel framework for stereo 3D object detection based on instance-level disparity estimation, which outperforms state-of-the-art baselines in terms of both accuracy and runtime speed.
 \item A pseudo-ground-truth generation process that provides supervision for the instance disparity estimation network and guides it to learn the object shape prior that benefits 3D object detection.
 \end{compactitem}
\section{Related Work}\label{sec:related-work}
In this section, we briefly review the recent progress of 3D object detection with different modalities of input data and introduce the background of object shape reconstruction that is used in the proposed pseudo-ground-truth generation process.

\medskip
\noindent\textbf{3D object detection with RGB images.}
Several works concentrate on 3D object detection using a monocular image or stereo RGB images as input.
Stereo R-CNN \cite{li2019stereo} designs a Stereo Region Proposal Network to match left and right Regions of Interest (RoIs), and refines 3D bounding boxes by dense alignment.
On the monocular side,
\cite{mousavian20173d} proposes to estimate 3D bounding boxes with relation and constraints between 2D and 3D bounding boxes.
\cite{xu2018multi} uses a depth map as an extra input channel to assist 3D object detection.
Recently, Pseudo-LiDAR \cite{wang2019pseudo} converts the disparity map estimated from stereo images to point clouds as pseudo-LiDAR points, estimates 3D bounding boxes with LiDAR-input approaches and achieves state-of-the-art performance on both monocular and stereo input.
It is worth noting that, there are two concurrent works OC-Stereo \cite{ponObjectCentricStereoMatching2019} and ZoomNet \cite{xu2020zoomnet} that propose the similar idea of instance-level disparity estimation. OC-Stereo \cite{ponObjectCentricStereoMatching2019} uses depth completion results from sparse LiDAR points as object-centric disparity supervision, and ZoomNet \cite{xu2020zoomnet} prepares a human-annotated CAD model dataset to achieve a similar purpose.
Our method differs from these above-mentioned works in the disparity estimation region (on objects vs. on full images) and the automated dense instance disparity pseudo-ground-truth generation process.

\medskip
\noindent\textbf{3D object detection with point clouds.}
A majority of state-of-the-art 3D object detection methods are based on point clouds captured by depth sensors (LiDAR or RGB-D camera) \cite{chen2017multi,qi2019deep} as input.
F-PointNet \cite{qi2018frustum} segments the object point cloud within the 2D RoI frustum into foreground and background and later predicts 3D bounding boxes with PointNet++ \cite{qi2017pointnet++}.
Recently, PointRCNN \cite{shi2019pointrcnn} adapts this framework into a two-stage design as in the 2D object detection counterpart \cite{ren2015faster} and achieved impressive performance.
The 3D object detector in the proposed pipeline is point cloud based and can be substituted to other methods that can achieve the similar purpose.

\begin{figure*}
    \centering
       \includegraphics[width=\linewidth]{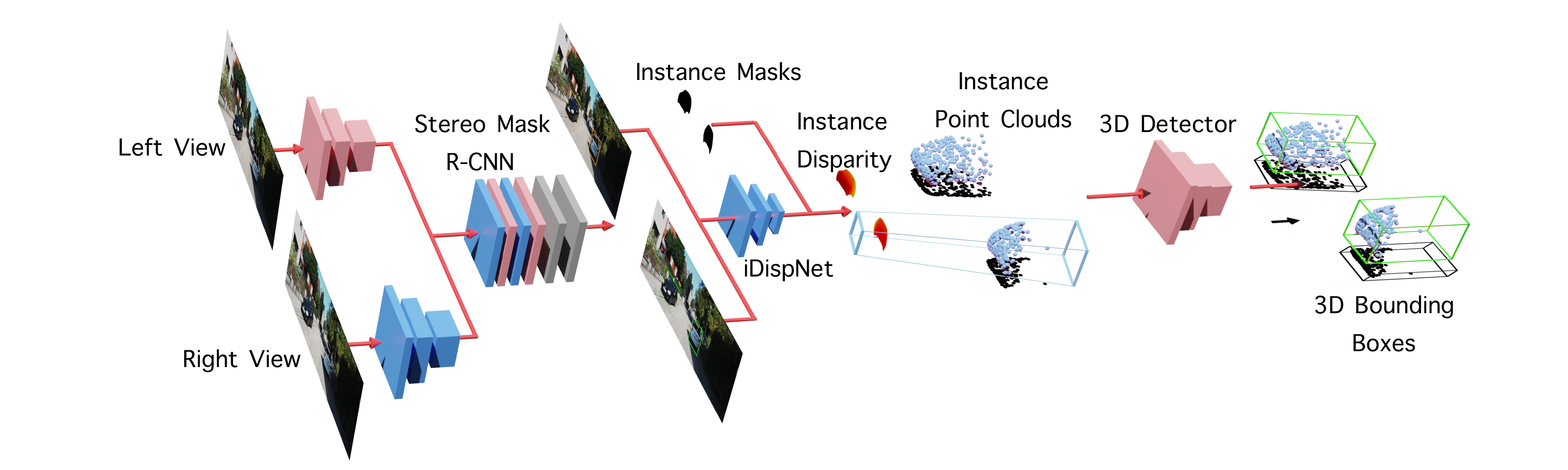}
       \caption{
           \textbf{Disp R-CNN Architecture.}
           Disp R-CNN has three stages. 
           First, the input images are passed through a stereo variant of Mask R-CNN to detect 2D bounding boxes and instance segmentation masks.
           Then, the instance disparity estimation network (iDispNet) takes the cropped RoI images as input and estimates an instance disparity map.
           Finally, the instance disparity map is converted to an instance point cloud and fed into the 3D detector for 3D bounding box regression.
           }
       \label{fig:arch}
   \end{figure*}

\medskip
\noindent\textbf{Object shape reconstruction.}
3D object detection can benefit from shape reconstruction.
\cite{engelmann2016joint} leverages the constraint that the point cloud must be lying on the object surface, and jointly optimizes the object pose and shape with the point cloud generated from stereo disparities and object shape prior model learned from the 3D shape repository with PCA.
\cite{murthy2017shape} further extents this pipeline with the temporal kinematic constraints of objects in dynamic scenes.
\cite{wang2020directshape} proposes a continuous optimization approach to jointly optimize object shape and pose with the photometric error.
\cite{manhardt2019roi} proposes to use the object shape generated from a 3D auto-encoder in the data augmentation process during the training of monocular 3D object detection.
For object categories other than vehicles (e.g., pedestrians and cyclists), shape reconstruction can be achieved similarly by fitting a statistical shape model (e.g., SMPL \cite{loper2015smpl}) to point cloud data as demonstrated by the PedX dataset \cite{kimPedXBenchmarkDataset2018}.

\section{Methods}
Given a pair of stereo images, the goal is to detect 3D bounding boxes of all the object instances of interest.
As shown in Fig. \ref{fig:arch}, our detection pipeline consists of three stages: we first detect 2D bounding boxes
and instance masks for each object, then estimate disparities only for pixels belonging to objects and finally use a 3D detector to predict 3D bounding boxes from the instance point cloud.

\subsection{Stereo Mask R-CNN}\label{sec:method-smrcnn}
We start by briefly describing the base 2D detector that provides necessary input for the following modules of the pipeline. We extend the Stereo R-CNN \cite{li2019stereo} framework to predict the instance segmentation mask in the left image. Stereo Mask R-CNN is composed of two stages.
The first stage is a stereo variant of the Region Proposal Network (RPN) as proposed in \cite{li2019stereo}, where object proposals from the left and right images are generated from the same set of anchors to ensure the correct correspondences between the left and right Regions of Interest (RoIs).
The second stage extracts object features from the feature map using RoIAlign as proposed in \cite{he2017mask}, followed by two prediction heads that produce 2D bounding boxes, classification scores, and instance segmentation masks.

\subsection{Instance Disparity Estimation Network}\label{sec:method-disp-est}
The disparity estimation module is responsible for recovering the 3D data in stereo 3D object detection and therefore its accuracy directly affects the 3D detection performance. 
Previous work \cite{wang2019pseudo} applies an off-the-shelf disparity estimation module that predicts the disparity map for all the pixels in the entire image.
Since the area of the foreground objects only takes a small portion of the full image, 
most computation in both the disparity estimation network and the object detection network is redundant and can be reduced.
Moreover, for the specular surfaces on most of the vehicles, the Lambertian reflectance assumption for the photometric-consistency constraint used in stereo matching cannot hold.
To remedy these problems, we propose a learning-based instance disparity estimation network (iDispNet) that is specialized for 3D object detection.
The iDispNet only takes the object RoI images as input and is only supervised on the foreground pixels, so that it captures the category-specific shape prior and thus produces more accurate disparity predictions.

\begin{figure}[tb]
    \centering
    \includegraphics[width = \linewidth]{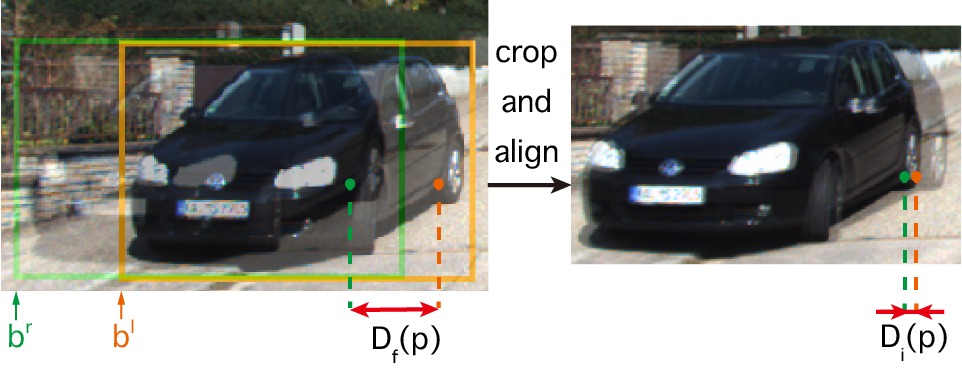}
    \caption{
        \textbf{The crop-and-align process} aligns the left and right RoIs by cutting off a \textit{global offset}.
        As a result, the instance disparity $D_i(p)$ distributes in a much narrower range compared to the full-frame disparity $D_f(p)$, which makes it possible to reduce the disparity search range when constructing the disparity cost volume and leads to faster inference.
    }
    \label{fig:crop_and_shift}
\end{figure}

Formally speaking, the \textit{full-frame disparity} for a pixel $p$ is defined as:
\begin{equation}
\label{eq:fullframedisp}
D_f(p)=u_p^l-u_p^r,
\end{equation}
where $u_p^l$ and $u_p^r$ represent the horizontal pixel coordinates of $p$ in the left and right views, respectively.
With the 2D bounding boxes produced by the Stereo Mask R-CNN, we can crop the left and right RoIs out from the full images and align them in the horizontal direction.
The width of each RoIs ($w^l$, $w^r$) are set to the larger value to make the two RoIs share the same size. 
Once RoIs are aligned, the disparity displacement for pixel $p$ on the left image (reference) changes from \textit{full-frame disparity} to 
\textit{instance disparity}, which is defined as:
\begin{equation}
\label{eq:instancedisp}
D_i(p)=D_f(p)-(b^l-b^r),
\end{equation}
where $b^l$ and $b^r$ stand for coordinates of the left border of bounding boxes in two views, respectively. %
Our goal is essentially to learn the instance disparity $D_i(p)$ instead of $D_f(p)$ for each $p$ belonging to an object of interest. This crop-and-align process is visually illustrated in Fig. \ref{fig:crop_and_shift}.

All the RoIs in the left and right images are resized to a common size $H\times W$.
For all the pixels ${p}$ that belong to an object instance $O$ given by the instance segmentation mask, 
the loss function for the instance disparities is defined as:
\begin{equation}
L_{idisp}=\dfrac{1}{|O|}\sum\limits_{p \in O}L_{1;smooth}(\hat D'_i(p)- D'_i(p)),
\end{equation}
\begin{equation}
D'_i(p)=\frac{D_i(p)}{\max(w^l, w^r)} W,
\end{equation}
where $\hat D'_i(p)$ is the predicted instance disparity for point $p$, $ D'_i(p)$ is the instance disparity ground-truth, 
$w^l$ and $w^r$ represent the widths of 2D bounding boxes in two views, and $|O|$ means the number of pixels belonging to the object $O$.

Once the iDispNet outputs instance disparity $\hat{D}_i'(p)$, we can compute the 3D location for each pixel $p$ belonging to the foreground as the input of the following 3D detector.
The 3D coordinate $(X,Y,Z)$ is derived as follows:
\begin{equation*}
    X=\dfrac{(u_p-c_u)}{f_u} Z, \quad
    Y=\dfrac{(v_p-c_v)}{f_v} Z,
\end{equation*}
\begin{equation*}
    Z=\dfrac{Bf_u}{\hat{D}_i(p)+b^l-b^r},
\end{equation*}
where $B$ is the baseline length between the left and right cameras,
$(c_u, c_v)$ is the pixel location corresponding to the camera center,
and $(f_u,f_v)$ are horizontal and vertical focal lengths, respectively.

\subsection{Pseudo Ground-truth Generation}\label{sec:method-pseudo-gt}

Training stereo matching network requires a large amount of dense disparity ground-truth, while most of the 3D object detection datasets \cite{geigerAreWeReady2012, caesar2019nuscenes, sun2019scalability} don't provide this data due to its difficulties in the manual annotation.
The full-frame disparity estimation module used in the recent works \cite{wang2019pseudo,you2019pseudo}
is first pre-trained on synthetic datasets and later fine-tuned on the real data with sparse disparity ground-truth converted from LiDAR points.
Although the detection performance gained large improvements from this supervision, the requirement for LiDAR point cloud limits the scaling capability of stereo 3D object detection methods in the real world scenario due to the high sensor price.

Benefiting from the design of the iDispNet which only requires foreground supervision, 
we propose an effective way to generate a large amount of dense disparity pseudo-ground-truth (pesudo-GT) for the real data without the need of LiDAR points.
The generation process is made possible by a category-specific shape prior model, from which the object shape can be reconstructed and later rendered to the image plane to obtain dense disparity ground-truth.

We use the volumetric Truncated Signed Distance Function (TSDF) as the shape representation.
For some rigid object categories with relatively small shape variations (e.g. vehicles), 
the TSDF shape space for this category can be approximated by a low-dimensional subspace \cite{leventon2002statistical, engelmann2016joint}.
Formally, denoting the basis of the subspace as $V$, which are obtained from the leading principal components of training shapes, and the mean shape as $\mu$, the shape $\tilde{\phi}$ of an instance can be represented as:
\begin{equation}
        \tilde{\phi}(z)=Vz+\mu,
\end{equation}
where $z\in \mathbb{R}^K$ is the shape coefficients and $K$ is the dimension of the subspace.

Given the 3D bounding box ground-truth and the point cloud of an instance, we can reconstruct shape coefficients $z$ for an instance by minimizing the following cost function:
\begin{equation}
        L_{pc}(z) = \dfrac{1}{|P|} \sum_{x\in P} \phi(x,z)^2,
\end{equation}
where $\phi(x,z)$ is the interpolated value of a 3D point $x$ in the TSDF volume defined by shape coefficients $z$, $P$ is the point cloud corresponding to the instance, and $|P|$ is the number of points in the point cloud.
Only $z$ is being updated through the optimization process.
Intuitively, this cost function minimizes the distance from the point cloud to the object surface defined by the zero crossing of the TSDF.
The point cloud can be obtained from an off-the-shelf disparity estimation module or optionally LiDAR points.

\begin{figure}[tb]
    \centering
    \includegraphics[width = 1.0\linewidth]{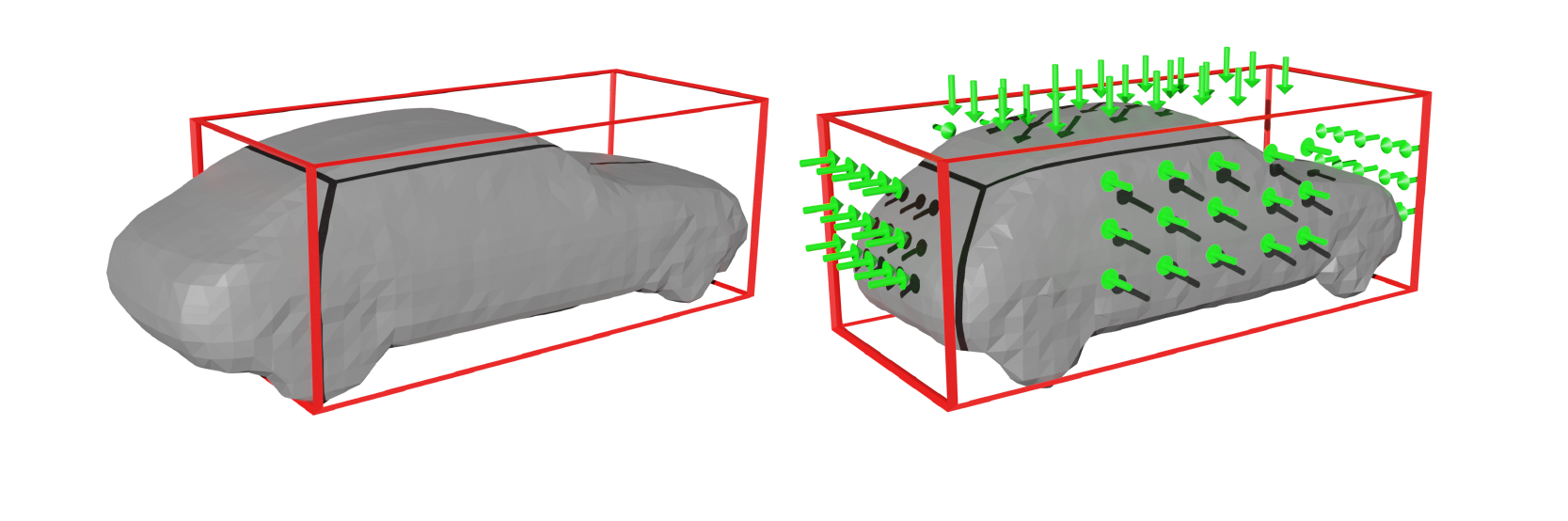}
    \caption{
        \textbf{The dimension regularization during Pseudo-GT generation}  penalizes a voxel if it is outside of the 3D bounding box and has a negative TSDF value,
        thus enforcing the shape surface to stay inside the 3D bounding box.
        From left to right: object shapes without and with dimension regularization.
    }
    \label{fig:dim-reg}
\end{figure}

Since the cost function above does not restrict the 3D dimension of object shape, we propose the following dimension regularization term to reduce the occurrence of objects overflowing the 3D bounding box:
\begin{equation}
        L_{dim}(z)=\sum_{v \in V^{out}} \max(-\phi(v,z),0)^2,
\end{equation}
where $V^{out}$ represents all the voxels that are defined outside of the 3D bounding box in a volume.
A visualization of the dimension regularization is shown in Fig. \ref{fig:dim-reg}.

To restrict the shape coefficients in an appropriate range, the following regularization term is used to penalize deviations of optimized shape from mean shape:
\begin{equation}
        L_z(z)=\sum_{k=1}^K (\frac{z_k}{\sigma _k})^2,
\end{equation}
where $\sigma_k$ is the $k$-th eigen value corresponding to the $k$-th principal component.

Combining the above terms, the total cost function is
\begin{equation}
        L(z)=w_1 L_{pc}(z)+w_2 L_{dim}(z)+w_3 L_z(z).
\end{equation}

Finally, instance disparity pseudo-GT $D_i$ can be rendered based on the optimized object shape as follows:
\begin{equation}
    D_i=\dfrac{Bf_u}{\pi(M(\tilde{\phi}(z)))}-(b^l-b^r),
\end{equation}
where $M$ represents the marching cubes \cite{lorensen1987marching} operation that converts the TSDF volume to a triangle mesh.
$\pi$ represents the mesh renderer that produces the pixel-wise depth map.
Some examples of the rendered disparity pseudo-GT are visualized in the third line of Fig. \ref{fig:qualitative}.

\subsection{Discussion}
\noindent\textbf{Choices on network design.}
There are two choices for the iDispNet design:
\textbf{(1)} Using only the decoder part of the iDispNet as a prediction head similar to the mask head in Mask R-CNN. 
The RoI feature extracted from the backbone is reused in disparity estimation and the disparity head is trained end-to-end with the rest of the network;
\textbf{(2)} Crop the RoI images from the original images, and then feed the cropped images to the encoder-decoder network of iDispNet.
As shown in the Table \ref{tab:disp_comparison} in the experiment section,
the result of \textbf{(1)} is suboptimal compared to \textbf{(2)}, 
so we choose \textbf{(2)} as the proposed design.
We believe the reason behind this result is related to the different requirements between the tasks of instance segmentation and disparity estimation.
Disparity estimation requires more fine-grained distinctive feature representation to make pixel-wise cost volume processing to be accurate, 
while instance segmentation is supervised to predict the same class probability for every pixel that belongs to the object.
By jointly training the end-to-end version of the network, the backbone has to balance between these two different tasks and thus causes the suboptimal result.

\medskip\noindent\textbf{Choices on the point cloud for Pseudo-GT generation.}
In general, there are two choices of point cloud usage in the shape optimization process.
The point cloud can be obtained from 
\textbf{(1)} the sparse LiDAR point clouds in the dataset with an optional depth completion step to improve density;
\textbf{(2)} the prediction of an off-the-shelf disparity estimation network trained on other datasets (e.g. PSMNet trained on KITTI Stereo).
\textbf{(1)} potentially gives a more accurate point cloud.
But for datasets or application scenarios without the LiDAR points as optimization target in $L_{pc}(z)$, \textbf{(2)} is the only choice.
We evaluate and present the results using both ways separately (titled by \textit{Ours (velo)} and \textit{Ours} relatively in Tab. \ref{tab:kitti_val} and \ref{tab:kitti_test}).
As later demonstrated in the results, \textbf{(2)} performs reasonably well without the usage of the LiDAR point cloud.

\subsection{Implementation Details}
\noindent\textbf{iDispNet.}\label{sec:imp-details-idispnet}
Following the setting in \cite{wang2019pseudo}, we use PSMNet \cite{changPyramidStereoMatching2018} as the architecture for iDispNet.
RoI images are cropped and resized to 224 $\times$ 224 as the input.
During stereo matching, we set the minimum and maximum instance disparities search range to -48 and 48 pixels, which cover 90\% of the cases according to the statistics for the disparity distribution across the training set.

\medskip\noindent\textbf{3D detection network.}
PointRCNN \cite{shi2019pointrcnn} is used as the 3D object detector in our implementation.
Different from inputting point clouds of the entire scene in the conventional approach, we use the instance point cloud converted from instance disparity as the input to PointRCNN.
The number of input point cloud subsamples is reduced to $768$.

\medskip\noindent\textbf{Pseudo-GT generation.}
To increase the stability of the pseudo-GT generation process, only points that sit inside of the ground-truth 3D bounding box are used for optimization. For objects with less than 10 points, the mean shape is directly used without further optimization.
Following \cite{engelmann2016joint}, we select the first five PCA components and set the volume dimension to $60\times40\times60$.
The training shapes are obtained from \cite{engelmann2016joint}, which are 3D models collected from the Google Warehouse website. 
During optimization, loss weights are set as $w_1=10/3, w_2=w_3=1$.
The optimization is achieved by a Levenberg–Marquardt solver implemented with Ceres \cite{ceres-solver}.

\begin{table*}
    \setlength{\belowcaptionskip}{-0.3cm}
    \begin{center}
    \renewcommand{\arraystretch}{1.3}
    \resizebox{\textwidth}{!}{
        \begin{tabular}{|c||c|c|c|c||c|c|c||c|c|c||c|c|c|} 
            \hline
            \multirow{2}{*}{Method} &
            \multirow{2}{*}{\makecell[c]{LiDAR \\ Supervision}} &
            \multicolumn{3}{c||}{$AP_{bev}$ (IoU=0.7) } & \multicolumn{3}{c||}{$AP_{3d}$ (IoU=0.7) }  & \multicolumn{3}{c||}{$AP_{bev}$ (IoU=0.5) }& \multicolumn{3}{c|}{$AP_{3d}$ (IoU=0.5) } \\ 
            \cline{3-14} & & Easy& Mod.& Hard& Easy& Mod.& Hard& Easy& Mod.& Hard& Easy& Mod.& Hard \\ 
            \hline\hline
            TL-Net \cite{qin2019triangulation} &  N & 29.22 & 21.88& 18.83& 18.15& 14.26& 13.72& 62.46& 45.99& 41.92 & 59.51 & 43.71 & 37.99\\ 
            \hline
            S-RCNN \cite{li2019stereo} & N & 68.50 & 48.30 & 41.47 & 54.11 & 36.69 & 31.07 & 87.13 & 74.11 & 58.93 & 85.84 & 66.28 & 57.24 \\ 
            \hline
            PL (AVOD) & N & 60.7 & 39.2 & 37.0 & 40.0 & 27.4 & 25.3 & 76.8 & 65.1 & 56.6 & 75.6 & 57.9 & 49.3\\ 
            \hline
            Ours & N & {\color{black}\textbf{76.51}} & {\color{black}\textbf{58.63}}  & {\color{black}\textbf{50.26}} & {\color{black}\textbf{63.57}} &  {\color{black}\textbf{47.15}} & {\color{black}\textbf{39.73}} & {\color{black}\textbf{90.60}} & {\color{black}\textbf{80.53}}  & {\color{black}\textbf{71.16}} & {\color{black}\textbf{90.38}} & {\color{black}\textbf{79.77}} & {\color{black}\textbf{69.81}} \\
            \hline\hline
            PL* (FP) & Y & 72.8 & 51.8 & 44.0 & 59.4 & 39.8 & 33.5 & 89.8 & 77.6 & 68.2 & 89.5 & 75.5 & 66.3 \\ 
            \hline
            PL* (AVOD)&  Y & 74.9 & 56.8 & 49.0 & 61.9 & 45.3 & 39.0 & 89.0 & 77.5 & 68.7 & 88.5 & 76.4 & 61.2\\ 
            \hline
            PL* (P-RCNN)&  Y  & 73.4 & 56.0 & {\color{black}\textbf{ 52.7 }}  & 62.3 & 44.9 & {\color{black}\textbf{41.6}}  & 88.4 & 76.6 & 69.0 & 88.0 & 73.7 & 67.8\\ 
            \hline
            Ours (velo)&  Y  & {\color{black}\textbf{77.63}}  & {\color{black}\textbf{64.38}}  & 50.68 & {\color{black}\textbf{64.29}}  & {\color{black}\textbf{47.73}}  & 40.11 & {\color{black}\textbf{90.67}}  & {\color{black}\textbf{80.45}}  & {\color{black}\textbf{71.03}}  & {\color{black}\textbf{90.47}}  & {\color{black}\textbf{79.76}}  & {\color{black}\textbf{69.71}}   \\ 
            \hline
            {\cellcolor[rgb]{0.902,0.902,0.902}} OC-Stereo& {\cellcolor[rgb]{0.902,0.902,0.902}} Y  &{\cellcolor[rgb]{0.902,0.902,0.902}} 77.66 &{\cellcolor[rgb]{0.902,0.902,0.902}} 65.95 &{\cellcolor[rgb]{0.902,0.902,0.902}} 51.20 &{\cellcolor[rgb]{0.902,0.902,0.902}} 64.07 &{\cellcolor[rgb]{0.902,0.902,0.902}} 48.34 &{\cellcolor[rgb]{0.902,0.902,0.902}} 40.39 &{\cellcolor[rgb]{0.902,0.902,0.902}} 90.01 &{\cellcolor[rgb]{0.902,0.902,0.902}} 80.63 & {\cellcolor[rgb]{0.902,0.902,0.902}}71.06 &{\cellcolor[rgb]{0.902,0.902,0.902}} 89.65 &{\cellcolor[rgb]{0.902,0.902,0.902}} 80.03 &{\cellcolor[rgb]{0.902,0.902,0.902}} 70.34 \\
            \hline
            {\cellcolor[rgb]{0.902,0.902,0.902}} ZoomNet&{\cellcolor[rgb]{0.902,0.902,0.902}}  - &{\cellcolor[rgb]{0.902,0.902,0.902}} 78.68 &{\cellcolor[rgb]{0.902,0.902,0.902}} 66.19 &{\cellcolor[rgb]{0.902,0.902,0.902}} 57.60 &{\cellcolor[rgb]{0.902,0.902,0.902}} 62.96 &{\cellcolor[rgb]{0.902,0.902,0.902}} 50.47 &{\cellcolor[rgb]{0.902,0.902,0.902}} 43.63 &{\cellcolor[rgb]{0.902,0.902,0.902}} 90.62 &{\cellcolor[rgb]{0.902,0.902,0.902}} 88.40 &{\cellcolor[rgb]{0.902,0.902,0.902}} 71.44 &{\cellcolor[rgb]{0.902,0.902,0.902}} 90.44 &{\cellcolor[rgb]{0.902,0.902,0.902}} 79.82 &{\cellcolor[rgb]{0.902,0.902,0.902}} 70.47 \\
            \hline
            {\cellcolor[rgb]{0.902,0.902,0.902}}PL++ (P-RCNN)&{\cellcolor[rgb]{0.902,0.902,0.902}}  Y  & {\cellcolor[rgb]{0.902,0.902,0.902}}82.0 &{\cellcolor[rgb]{0.902,0.902,0.902}} 64.0 & {\cellcolor[rgb]{0.902,0.902,0.902}}57.3  &{\cellcolor[rgb]{0.902,0.902,0.902}} 67.9 &{\cellcolor[rgb]{0.902,0.902,0.902}} 50.1 & {\cellcolor[rgb]{0.902,0.902,0.902}}45.3  &{\cellcolor[rgb]{0.902,0.902,0.902}} 89.8 & {\cellcolor[rgb]{0.902,0.902,0.902}}83.8 & {\cellcolor[rgb]{0.902,0.902,0.902}}77.5 &{\cellcolor[rgb]{0.902,0.902,0.902}} 89.7 &{\cellcolor[rgb]{0.902,0.902,0.902}} 78.6  &{\cellcolor[rgb]{0.902,0.902,0.902}} 75.1 \\
            \hline
            \end{tabular}
    }
    \end{center}
    \caption{
    \textbf{3D object detection results on the KITTI object validation set. }
    We report average precision of bird's eye view (AP$_{\rm bev}$) and 3D boxes (AP$_{\rm 3d}$) for the \textbf{car} category.
    LiDAR supervision indicates if the method uses the sparse LiDAR point cloud as a supervision signal during training.
    We report the reproduced result for PL (AVOD) since \cite{wang2019pseudo} didn't provide full results on experiments without LiDAR supervision.
    Besides published state-of-the-art methods, we also present the results of concurrent works (grey background) for comparison.
    }
    
    \vspace{-0.3cm}
    \label{tab:kitti_val}
    \end{table*}
\medskip\noindent\textbf{Training strategy.}
We train the Stereo Mask R-CNN for $20$ epochs with a weight decay of $0.0005$, the iDispNet for $100$ epochs with a weight decay of $0.01$ and the PointRCNN $360$ epochs with a weight decay of $0.0005$.
The learning rate is first warmed up to $0.01$ and then decreases slowly in all the training processes.

\section{Experiments}\label{sec:exp}
\begin{table}
\begin{center}
\renewcommand{\arraystretch}{1.3}
\resizebox{0.47\textwidth}{!}{
    \begin{tabular}{|c||c|c|c||c|c|c|} 
        \hline
        \multirow{2}{*}{Method}&\multicolumn{3}{c|}{$AP_{bev}$ (IoU=0.7)}  & \multicolumn{3}{c|}{$AP_{3d}$ (IoU=0.7) } \\ 
        \cline{2-7} & Easy    & Mod.    & Hard    & Easy  & Mod.   & Hard     \\ 
        \hline\hline
        S-RCNN & 61.67& 43.87& 36.44  & 49.23& 34.05& 28.39 \\ 
        \hline
        PL* (FP)& 55.0& 38.7& 32.9& 39.7& 26.7& 22.3\\ 
        \hline
        PL* (AVOD)& 66.83& 47.20& 40.30& 55.40& 37.17& 31.37\\ 
        \hline
        Ours & 73.82 & \textbf{52.34} & 43.64& 58.53 & 37.91& 31.93\\
        \hline
        Ours (velo) & \textbf{74.07} & \textbf{52.34} & \textbf{43.77}& \textbf{59.58}& \textbf{39.34} & \textbf{31.99}\\ 
        \hline
        {\cellcolor[rgb]{0.902,0.902,0.902}}ZoomNet & {\cellcolor[rgb]{0.902,0.902,0.902}}72.94 & {\cellcolor[rgb]{0.902,0.902,0.902}}54.91 & {\cellcolor[rgb]{0.902,0.902,0.902}}44.14 & {\cellcolor[rgb]{0.902,0.902,0.902}}55.98 & {\cellcolor[rgb]{0.902,0.902,0.902}}38.64 & {\cellcolor[rgb]{0.902,0.902,0.902}}30.97                           \\ 
        \hline
        {\cellcolor[rgb]{0.902,0.902,0.902}}OC-Stereo      & {\cellcolor[rgb]{0.902,0.902,0.902}}68.89 & {\cellcolor[rgb]{0.902,0.902,0.902}}51.47 & {\cellcolor[rgb]{0.902,0.902,0.902}}42.97 & {\cellcolor[rgb]{0.902,0.902,0.902}}55.15 & {\cellcolor[rgb]{0.902,0.902,0.902}}37.60 & {\cellcolor[rgb]{0.902,0.902,0.902}}30.25                   \\ 
        \hline
        {\cellcolor[rgb]{0.902,0.902,0.902}}PL++    & {\cellcolor[rgb]{0.902,0.902,0.902}}75.5  & {\cellcolor[rgb]{0.902,0.902,0.902}}57.2  & {\cellcolor[rgb]{0.902,0.902,0.902}}53.4  & {\cellcolor[rgb]{0.902,0.902,0.902}}60.4  & {\cellcolor[rgb]{0.902,0.902,0.902}}44.6  & {\cellcolor[rgb]{0.902,0.902,0.902}}38.5                       \\
        \hline
        \end{tabular}
}
\end{center}
\caption{\textbf{3D object detection results on the KITTI object test set. }
We report Average Precision of bird's eye view (AP$_{\rm bev}$) and 3D boxes (AP$_{\rm 3d}$) for \textbf{car} category.
\textit{Ours (velo)} and \textit{Ours} indicates our method that uses and does not uses the sparse LiDAR point cloud as a supervision, respectively.
Besides published state-of-the-art methods, we also present the results of concurrent works (grey background) for comparison. }
\vspace{-0.3cm}
\label{tab:kitti_test}
\end{table}
We evaluate the proposed approach on the 3D object detection benchmark of KITTI dataset \cite{geigerAreWeReady2012}.
First, we compare our method to state-of-the-art methods on the KITTI object detection benchmark in Sec. \ref{sec:exp-3ddet-results}.
Next, we conduct ablation studies to analyze the effectiveness of different components of the proposed method in Sec. \ref{sec:exp-ablation}.
Then, we report the running time of our method in Sec. \ref{sec:exp-running-time}.
Finally, we provide some failure cases of our method in Sec. \ref{sec:exp-failure-cases}.

\subsection{3D Object Detection on KITTI}\label{sec:exp-3ddet-results}
The KITTI object detection benchmark contains 7481 training images and 7518 testing images.
To evaluate on the training set, we divide it into the training split and the validation split with 3712 and 3769 images following \cite{chen20153d}, respectively.
Objects are divided into three levels: easy, moderate and hard, depending on their 2D bounding box sizes, occlusion, and truncation extent following the KITTI settings.

\medskip\noindent\textbf{Evaluation of 3D object detection.}
We evaluate our method and compare it to previous state-of-the-art methods on the KITTI object 3D detection benchmark \cite{geigerAreWeReady2012}.
We perform the evaluation using Average Precision (AP) for 3D detection and bird's eye view detection.

In Tab. \ref{tab:kitti_val}, we compare our method with previous state-of-the-art methods on the validation split using 0.7 and 0.5 as the IoU threshold.

PL \cite{wang2019pseudo} estimates full-frame disparities, while our iDispNet predicts disparities only for pixels on objects.
When LiDAR supervision is not used at training time, our method outperforms PL (AVOD) over 10\% AP in all metrics.
Specifically, our method gains over 23.57\% improvement for $AP_{bev}$ in the easy level with an IoU threshold of 0.7.
This huge improvement comes from the pseudo-GT generation, which can provide a large amount of training data even if LiDAR ground-truth is not available at training time.

When LiDAR supervision is used at training time, our method still outperforms previous state-of-the-art methods in most of the metrics.
PL* (P-RCNN) and ours share the same 3D detector, but our method still obtains better results.
Specifically, our method gains an 8.38\% improvement in $AP_{bev}$ at the moderate level with an IoU threshold of 0.7.
The reason is that our iDispNet focuses on the foreground regions and we have much denser training data via the object shape rendering.

\begin{table}
\setlength{\belowcaptionskip}{-0.3cm}
\begin{center}
    \renewcommand{\arraystretch}{1.3}
    \resizebox{0.47\textwidth}{!}{
        \begin{tabular}{|c||c|c|c||c|c|} 
            \hline
            \multirow{2}{*}{Method} &
            \multirow{2}{*}{GT} &\multicolumn{2}{c||}{Pixel-wise}   & \multicolumn{2}{c|}{Object-wise}    \\ 
            \cline{3-6} & & Disparity     & Depth         & Disparity     & Depth          \\ 
            \hline\hline
            PSMNet & PGT&  1.53 & 0.54 & 0.87 & 1.00 \\ 
            \hline
            Ours (e2e) & PGT & 1.22 & 0.41 & 0.76 & 0.86 \\ 
            \hline
            Ours &PGT&  \textbf{0.90} & \textbf{0.28} & \textbf{0.38} & \textbf{0.33}  \\
            \hline \hline
            PSMNet &LiDAR&  1.01 & 0.64 & 1.27 & 1.28  \\
             \hline
            GANet &LiDAR&  \textbf{0.89} & 0.63 & \textbf{1.23} & 1.24  \\
             \hline
            Ours &LiDAR& 1.32 & \textbf{0.60} & 1.27 & \textbf{1.06}  \\
            \hline
            \end{tabular}
    }
\end{center}
\caption{\textbf{Disparity EPE and Depth RMSE comparison}, evaluated on the KITTI \textit{validation set}. 
We use our disparity pseudo-GT and sparse LiDAR as ground-truth for evaluation, denoted by PGT and LiDAR respectively.}
\vspace{-0.3cm}
\label{tab:disp_comparison}
\end{table}

Tab. \ref{tab:kitti_test} compares our method with previous state-of-the-art methods and several concurrent works on the KITTI test set with an IoU threshold of 0.7.
Comparing with previous methods, our method achieves the state-of-the-art performance in all metrics.
Specifically, our method gains 7\% and 5\% improvement in $AP_{bev}$ at the easy and moderate levels, respectively, and 4\% improvement in $AP_{3d}$ at the easy level, comparing to the previous state-of-art PL* (AVOD).
Among concurrent works, OC-Stereo \cite{ponObjectCentricStereoMatching2019} and ZoomNet \cite{xu2020zoomnet} share a similar idea with ours.
OC-Stereo utilizes LiDAR points after completion as supervision, and ZoomNet introduces fine-grained annotations to generate the ground-truth.
Instead, our pseudo-GT is rendered from the optimized object shape, which is more accurate than OC-Stereo and more efficient than ZoomNet, and thus leads to better performance on the KITTI test set.
More remarkably, our method achieves the state-of-the-art performance even if LiDAR supervision is not used at training time, which further shows that our method is robust and applicable in real-world applications.

\begin{figure*}
\vspace{-3em}
\centering
\includegraphics[width=\linewidth]{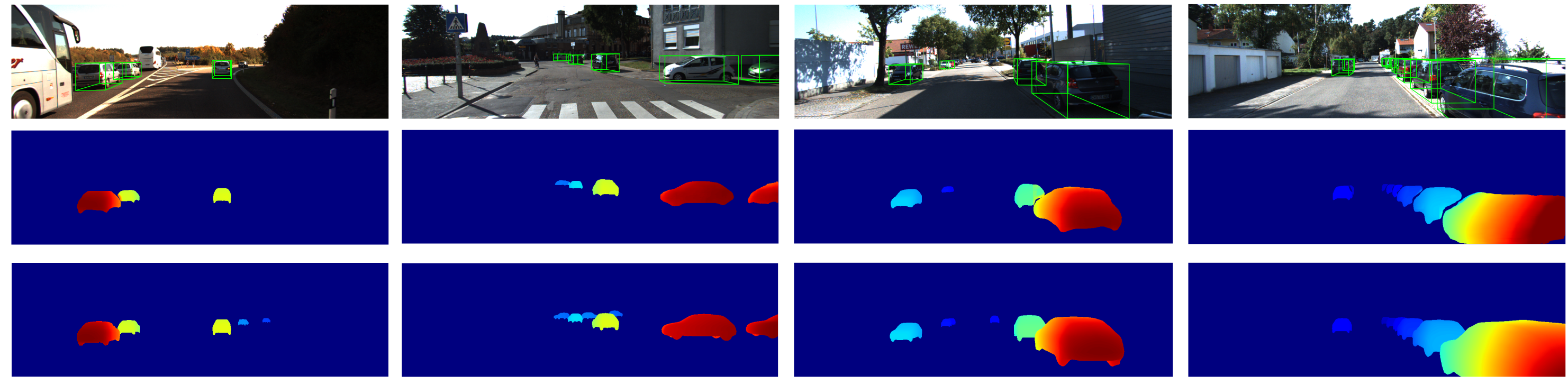}
\caption{\textbf{Qualitative results.} The rows from top to bottom present 3D bounding box prediction, instance disparity estimation and our disparity pseudo-ground-truth, respectively.}
\label{fig:qualitative}
\end{figure*}

We visualize some qualitative results of object detection, instance disparity estimation, and disparity pseudo-GT in Fig. \ref{fig:qualitative}.

\subsection{Ablation Studies}\label{sec:exp-ablation}
In this section, we conduct extensive ablation experiments to analyze the effectiveness of different components in our method.

\medskip\noindent\textbf{Cost function for shape optimization.}
To measure the effectiveness of the dimension regularization in the shape optimization process, we perform optimization processes with and without dimension regularization, and then compute the percentage of objects that have more than 70\% vertices locating inside the 3D bounding box. Our experiments show that the use of dimension regularization makes the percentage above rise from 71\% to 82\%, which proves that considering dimension regularization can reduce the occurrence of shape overflowing the 3D bounding box, thereby improving the quality of the object shape and the pseudo-GT.

\medskip\noindent\textbf{Instance disparity estimation.}
To validate the benefit of instance disparity estimation, we compute the disparity end-point-error (EPE) and depth RMSE for our iDispNet and some full-frame deep stereo networks in the foreground area.

In addition to the pixel-wise error, we also calculate the object-wise error, which is defined as the average error within each instance, and then averaged among instances.
We believe that the object-wise error is more suitable to reflect the quality of disparity estimation for each object because the pixel-wise error is dominated by objects with large areas.

The results are in Tab. \ref{tab:disp_comparison}. We use the pseudo-GT and sparse LiDAR as ground-truth separately, denoted by PGT and LiDAR.
PSMNet and GANet are trained on the KITTI Stereo dataset, while our iDispNet is trained with the pseudo-GT.
With the pseudo-GT as ground-truth, our iDispNet reaches smaller disparity and depth errors than the full-frame PSMNet by a large margin.
With sparse LiDAR points as ground-truth, our iDispNet still performs better than the full-frame method PSMNet and the state-of-the-art deep stereo method GA-Net \cite{zhang2019ga}, especially for the object-wise depth RMSE error.

Comparing the second and third lines in Tab. \ref{tab:disp_comparison} shows that re-using the features extracted from the RPN limits the quality of estimated disparity maps, which leading the end-to-end version of the iDispNet to give sub-optimal results, so we don't report results of the end-to-end version in other experiments.

\begin{figure}
{\centering
\resizebox{0.45\textwidth}{!}{
\includegraphics[width=15cm]{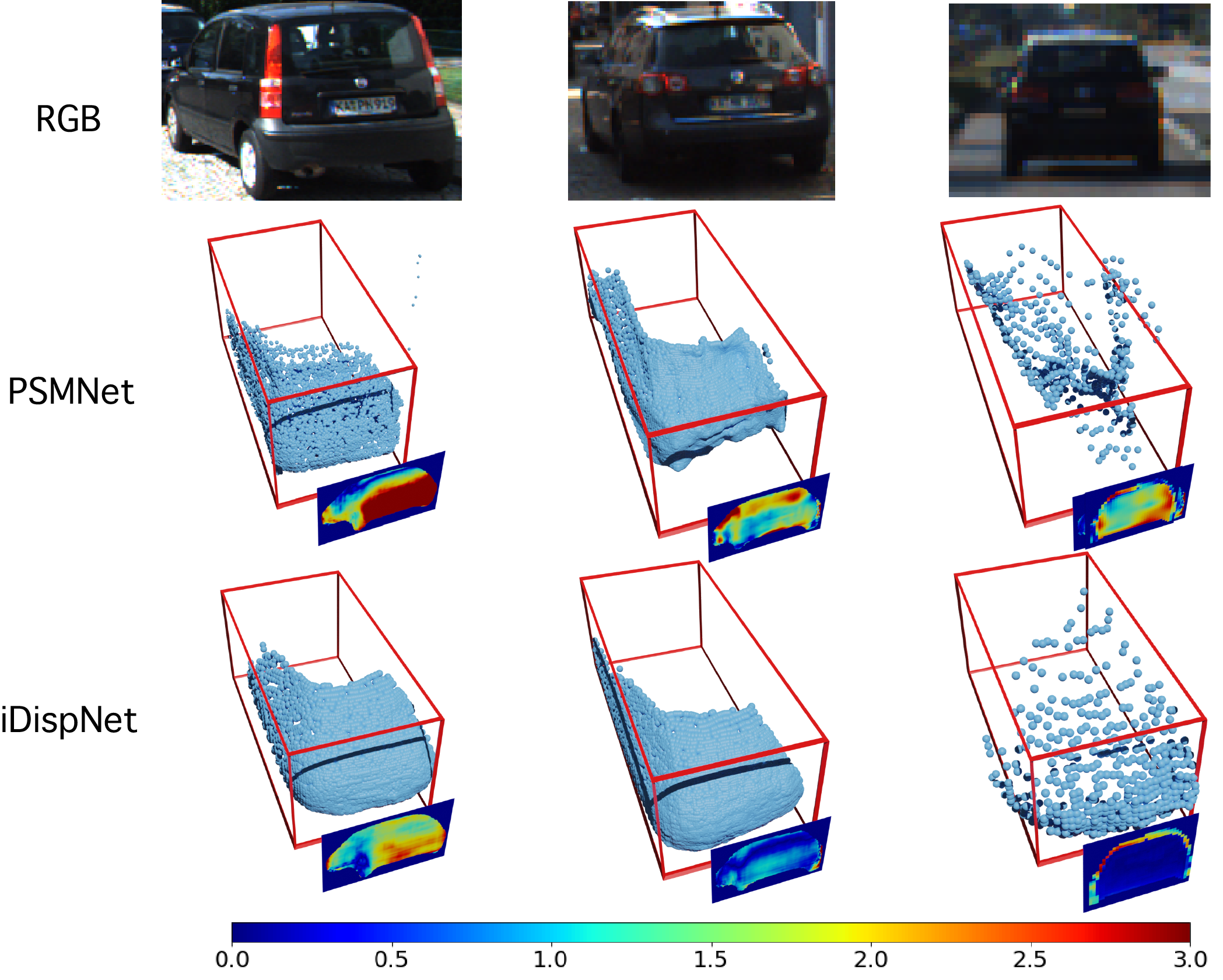}}
}
\vspace{0.5em}
\caption{\textbf{Qualitative comparison of disparity estimation results between PSMNet and our iDispNet.} 3D ground-truth bounding boxes are shown in red. Disparity error maps are shown as well, where the larger value indicates the worse disparity.}
\label{fig:disp_quality}
\end{figure}

Some qualitative results of instance disparity estimation and the comparison against the full-frame disparity estimation are shown in Fig. \ref{fig:disp_quality}.
The full-frame PSMNet cannot capture the smooth surfaces and sharp edges of vehicles, thus leading the following 3D detector to struggle to predict correct bounding boxes from inaccurate point clouds.
In contrast, our iDispNet gives more accurate and stable predictions thanks to instance disparity estimation and the supervision from the disparity pseudo-GT.
\subsection{Running Time}\label{sec:exp-running-time}
\begin{table}
\setlength{\belowcaptionskip}{-0.3cm}
\begin{center}
\renewcommand{\arraystretch}{1.3}
\resizebox{0.46\textwidth}{!}{
    \begin{tabular}{|l||c|c|c|c|c|} 
        \hline
        Method   & S-RCNN & PL (AVOD) & PL (PRCNN) & PL (FP) & Ours \\ 
        \hhline{|=::=====|}
        Time (s) & 0.417  & 0.51      & 0.51       & 0.67    & 0.425 \\
        \hline
        \end{tabular}
}
\end{center}
\vspace{-0.3 mm}
\caption{\textbf{Running time comparison.} S-RCNN represents Stereo R-CNN \cite{li2019stereo}.}
\label{tab:run-time}
\end{table}

Tab. \ref{tab:run-time} shows the running time comparison of our method and other stereo methods.
Our method takes 0.425s at inference time, surpassing almost all prior stereo methods.
Specifically, our method takes 0.17s for the 2D detection and segmentation, 0.13s for the instance disparity estimation, and 0.125s for the 3D detection from the point cloud.
The efficiency is attributed to estimating only the disparity in RoIs and only the 3D bounding boxes from the instance point clouds, which greatly reduces the search space.

\subsection{Failure Cases}\label{sec:exp-failure-cases}
\begin{figure}
    \centering
    \resizebox{\columnwidth}{!}{
        \begin{tabular}{ccc}
            \includegraphics[width=0.33\linewidth]{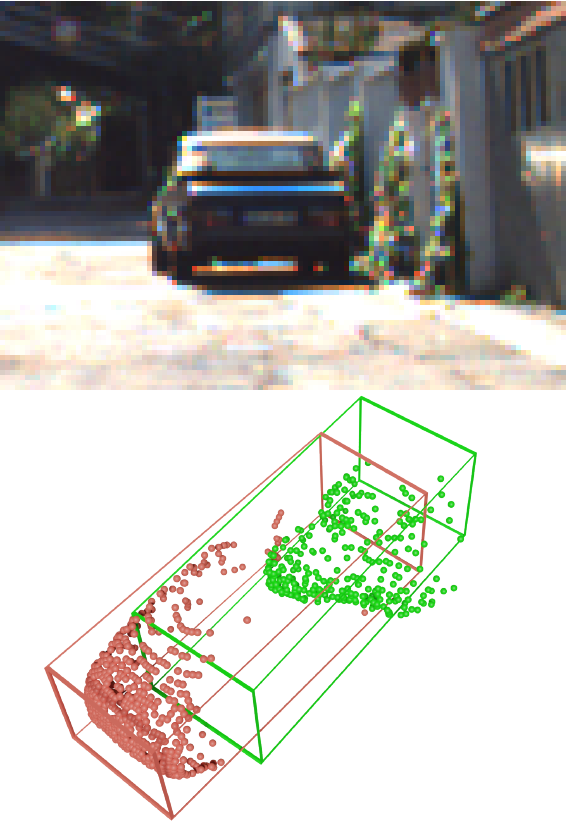} &
            \includegraphics[width=0.33\linewidth]{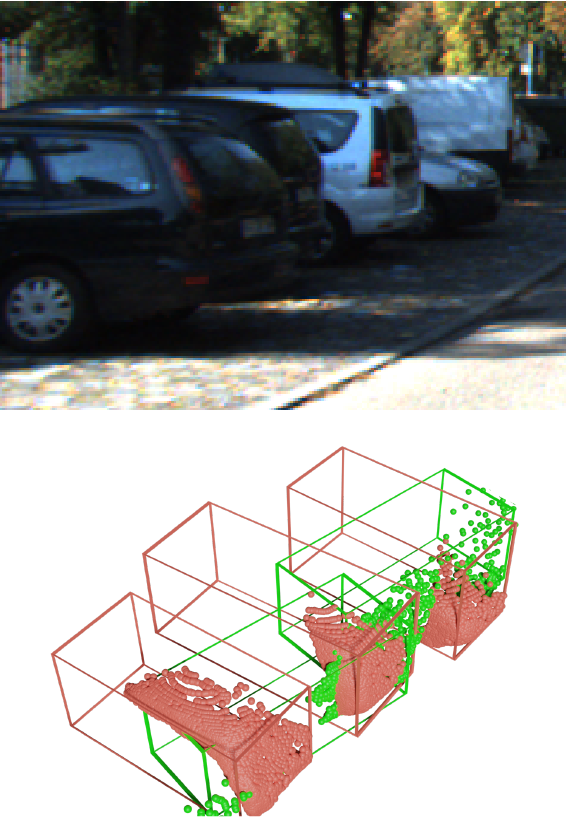}&
            \includegraphics[width=0.33\linewidth]{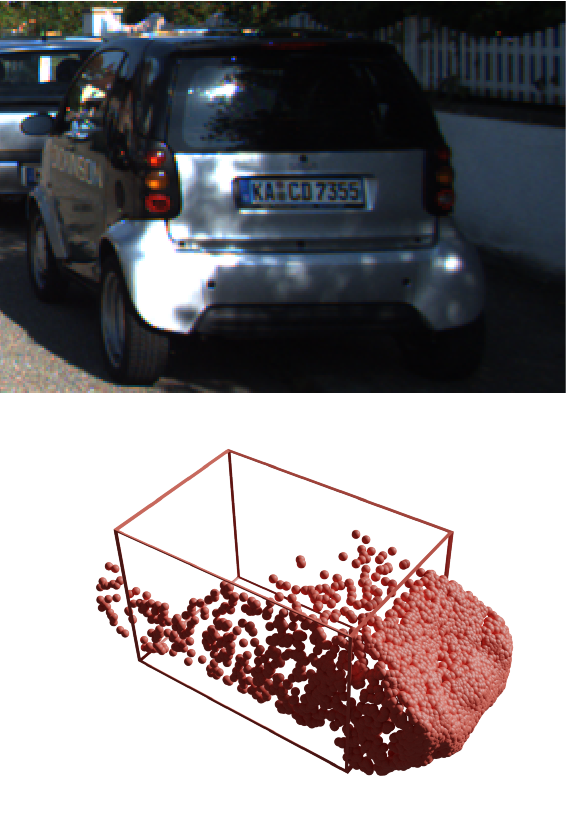}\\
            (a) & (b) & (c) 
          \end{tabular}
        
    }
    \centering
    \vspace{0.5em}
    \caption{\textbf{Failure cases.} The ground-truth bounding boxes and the pseudo-GT point clouds are visualized in {\color[rgb]{0.98,0.5,0.447}red}, while the predictions are visualized in {\color{green} green}.}
    \label{fig:failurecase}
\end{figure}

We visualize some failure cases in Fig. \ref{fig:failurecase}.
Our 3D object detection method is most likely to fail on objects that are too far away as shown in Fig. \ref{fig:failurecase}(a), or under strong occlusion or truncation as shown in Fig. \ref{fig:failurecase}(b). 
The reason is that there are too few 3D points on these objects for the detector to predict the correct bounding boxes.
Our pseudo-GT generation is most likely to fail on objects with unusual shapes, such as the car in Fig. \ref{fig:failurecase}(c) which is much shorter than other cars. Since there are very few examples with this kind of shape in the CAD model training set, so it is difficult to reconstruct this type of cars with the statistical shape model.

\section{Conclusion}
In this paper, we proposed a novel approach for 3D object detection from stereo images.
The key idea is to estimate instance-level pixel-wise disparities only in detected 2D bounding boxes and detect objects based on the instance point clouds converted from the instance disparities.
To solve the scarcity and sparsity of the training data, we proposed to integrate shape prior learned from CAD models to generate pseudo-GT disparity as supervision.
Experiments on the 3D detection benchmark of the KITTI dataset showed that our proposed method outperformed state-of-the-art methods by a large margin, especially when LiDAR supervision was not available at training time. We believe that the proposed approach is also applicable to other object categories, e.g., pedestrians and cyclists, whose shapes can be reconstructed similarly by fitting a statistical shape model (e.g., SMPL \cite{loper2015smpl}) to point cloud data, as demonstrated by the PedX dataset \cite{kimPedXBenchmarkDataset2018}.

\vspace{1em}
\noindent\textbf{Acknowledgements:} 
The authors would like to acknowledge support from NSFC (No. 61806176), Fundamental Research Funds for the Central Universities (2019XZZX004-09) and ZJU-SenseTime Joint Lab of 3D Vision.

{\small
\bibliographystyle{ieee_fullname}
\normalem
\bibliography{drcnn_paper}
}

\end{document}